\documentclass{article}

\usepackage{amssymb,amsmath}
\usepackage{graphicx}
\usepackage{subfigure}
\usepackage{dcolumn}
\usepackage[latin1]{inputenc}
\usepackage{url}
\urldef{\mailsa}\path|{brandoli}@icmc.usp.br|
\urldef{\mailsa}\path|{wnunes}@ursa.ifsc.usp.br|
\urldef{\mailsc}\path|{bruno}@ifsc.usp.br|

\begin{document}

\title{Material quality assessment of silk nanofibers based on swarm intelligence}

\author{
Bruno Brandoli Machado $^{1,3}$
 \and 
 Wesley Nunes Gon\c{c}alves $^{2,3}$ 
 \and 
 Odemir Martinez Bruno $^{1,2}$ 
\\
\\
$^1$ University of S\~ao Paulo (USP) - Brazil\\Institute of Mathematical Sciences and Computing\\
\\
$^2$  University of S\~ao Paulo (USP) - Brazil\\S\~ao Carlos Institute of Physics\\ 
\\
$^3$ Federal University of Mato Grosso do Sul - Brazil\\
\\
\small{Contacts:}\\ 
\small{brandoli@icmc.usp.br}\\ 
\small{wnunes@ursa.ifsc.usp.br}\\
\small{bruno@ifsc.usp.br}
}

\maketitle

\begin{abstract}
In this paper, we propose a novel approach for texture analysis based on artificial crawler model. Our method assumes that each agent can interact with the environment and each other. The evolution process converges to an equilibrium state according to the set of rules. For each textured image, the feature vector is composed by signatures of the live agents curve at each time. Experimental results revealed that combining the minimum and maximum signatures into one increase the classification rate. In addition, we pioneer the use of autonomous agents for characterizing silk fibroin scaffolds. The results strongly suggest that our approach can be successfully employed for texture analysis.
\\
\\
\textit{keywords: agent-based model, texture analysis, silk fibroin scaffolds}
\end{abstract}

\section{Introduction}

The silk fibroin is a protein extracted from cocoons of \textit{Bombyx mori} silkworm. It has been widely used in biomedical applications due to its high capacity to suture tissues, to regenerate bones and its biocompatibility with several types of human cells used in prosthesis~\cite{altamnBIOMATERIALS2003,luBIOMACROMOLECULES2010}. In the last years, researchers have proposed to improve the nanoscale features of silk fibroin by adding glycerol~\cite{luBIOMACROMOLECULES2010}. Though the presence of glycerol can provide better material properties, it also can alter the silk fibroin molecules interactions, damaging the result in its surface. Therefore, texture analysis methods emerge as a powerful tool for determining the suitable concentration of glycerol.

Many methods for texture description have been proposed in the literature. These methods are based on \textit{statistical analysis} of the spatial distribution (e.g., co-occurrence matrices~\cite{haralickTSMC1973}, local binary pattern~\cite{kashyapPAMI1986} and entropy~\cite{Fabbri20124487}), \textit{stochastic models} (e.g., Markov random fields~\cite{crossPAMI1983}), \textit{spectral analysis} (e.g., Fourier descriptors~\cite{azencottPAMI1997}, Gabor filters~\cite{machado2011,gaborJIEE1946} and wavelets transform~\cite{daubechies1992}), \textit{complexity analysis} (e.g., fractal dimension~\cite{odemirIS2008}), \textit{agent-based model} (e.g., deterministic tourist walk~\cite{backesPR2010,goncalvesESWA2012}). Although there are effective texture methods, they do not capture the richness of patterns of the silk fibroin scaffolds.

In this paper, we present a methodology for classifying surface properties of silk fibroin by means of texture analysis. The texture description approach proposed here is based on the artificial crawler model~\cite{zhangIAT2004,zhangIJPRAI2005}. We propose a new rule of movement that not only moves artificial crawler agents toward higher intensity, as well as to lower ones. We confirm that this strategy increases the discriminatory power and outperforms the state-of-the-art method.

This paper is organized as follows. Section \ref{sec:acrawler} details the original artificial crawler model. Section \ref{sec:approach} presents our approach to characterize textured images. Section \ref{sec:exps} discusses the results of the experiments. Finally, conclusions are given in Section \ref{sec:conclusions}.

\section{The Original Artificial Crawler Model} 
\label{sec:acrawler}

The first artificial crawler (ACrawler) model was developed in \cite{zhangIAT2004,zhangIJPRAI2005}. Let
us consider that an image is a pair $(\Upsilon,I)$, consisting of a finite set $\Upsilon$ of pixels, and a mapping $I$
that assigns to each pixel $p = (x_p,y_p) \in \Upsilon$ an intensity $I(p)$ ranging from $0$ to $255$. A pixel of the
intensity map $I$ holds a neighborhood set $\eta(p)$ of pixels $q$, which $d(p,q) \leq \sqrt{2}$ is the Euclidean
distance between pixels $p$ and $q$. Thereby, we assume the eight-connected neighbors.

\begin{figure}[!htb]
	\centering
		\subfigure[]{\includegraphics[width=0.29\textwidth]{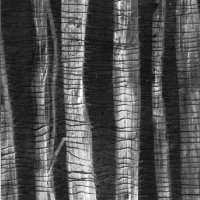}}
		\subfigure[]{\includegraphics[width=0.283\textwidth]{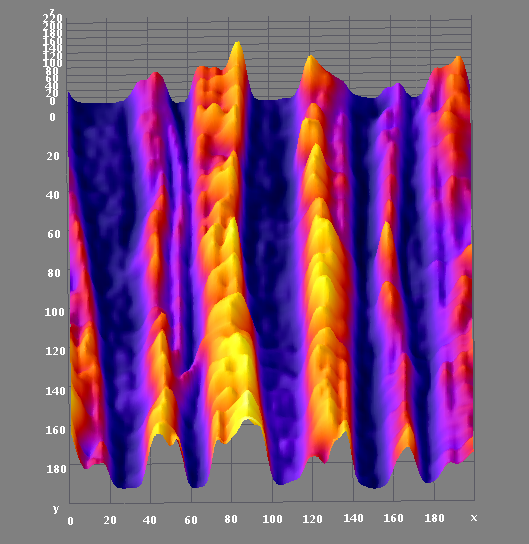}}
	\caption{\label{pic:environment} The environment of the agents. On the most left is shown a textured image (a)
and its respective 3D surface (b).}
\end{figure}

The original artificial crawler model assumes that each agent lives on one pixel of the image. At each time $t$, an
agent $i$ is characterized by two attributes: (1) $e_{i}^{t}$ holds a level of energy and (2) $\rho_{i}^{t}$ occupies a
spatial position in the image. First, $n$ agents are born with identical energy $\epsilon$. Such energy
can either wax or wane their lifespan according to energy consumption and influence of the environment. On images, the
environment is treated as a 3D surface with different altitudes that correspond to grey values in z-axis of the images.
Higher intensities supply nutrients to the agents, while lower altitudes correspond to the land. Figure \ref{pic:environment} shows a textured image and the peaks and valleys where the agents can live. 

The algorithm consists of a set of rules that comprises the evolution process:

\begin{enumerate}
 \item \textbf{Born}: Each agent $i$ is born with the same energy $\epsilon$.
$$\forall_{i},e_{i}^{0} = \epsilon$$

 \item \textbf{Survival threshold}: An agent $i$ dies if its energy is below the threshold.
$$\forall_{t,i}, \textrm{ if } e_{i}^{t} \leq e_{min} \textrm{ then } i \textrm{ dies }$$

 \item \textbf{Movement}: $$\forall_{i}: e_{i}^{t} > e_{min}, \rho_{i}^{t+1} = f(\rho_{i}^{t})$$

$$f(\rho) =
 \begin{cases}
 \rho_{i}^{t}, & \text{if (a) is satisfied} \\
 \rho_{max}^{t}, & \text{if (b) is satisfied} \\
 \rho_{m}^{t}, & \text{if (c) is satisfied}
 \end{cases}$$ \\
$(a)$ Agents settle down if the grey level of its 8-neighbors are lower than itself. \\
$(b)$ Agents move to a specific pixel if there exist one of its 8-neighbors ($\rho_{max}^{t}$) with higher intensity \\
$(c)$ If there exist more than one neighbor with higher intensity, an agent moves to the pixel that already was
occupied ($\rho_{m}^{t}$). \\

 \item \textbf{Energy consumption}: Each time $t$ consumes one unity of energy.
$$\forall_{i}: e_{i}^{t} > e_{min}, e_{i}^{t+1} = e_{i}^{t} - e_{unity}$$

 \item \textbf{Law of the jungle}: An agent with higher energy eats up another with lower one.
$$\forall_{i,j}: \rho_{i}^{t+1} = \rho_{j}^{t+1},e_{max}\{e_{i}^{t},e_{j}^{t}\} = \max\{e_{i}^{t},e_{j}^{t}\}$$

 \item \textbf{Gain of energy}: It up to dates the energy absorption from the environment, where $\lambda$ is a rate of
absorption over the pixel $I(\rho_i^{t})$.
$$\forall_{i}, e_{i}^{t+1} = e_{i}^{t} + \lambda I(\rho_i^{t})$$

 \item \textbf{Limit of energy}: It bounds the maximum energy $e_{max}$.
$$\forall_{t,j}: e_{j}^{t} \geq e_{max}, e_{j}^{t+1} = e_{max}$$
\end{enumerate}

Agents that were born in lower altitudes areas can die in the evolution process, while individuals that reached to
settle down in areas of higher altitudes have higher likelihood to remain alive.
To quantify the multi-agent system, a curve of live agents at each time is obtained:

\begin{equation}
\varphi = [\psi(0), \psi(1), \dots, \psi(t_{max})]
\end{equation}
where $\psi(t)$ is the number of live agents at time $t$ and $t_{max}$ is maximum iteration.

\section{A Novel Approach with Artificial Crawler to Texture Analysis}
\label{sec:approach}

The artificial crawler model described above consists of moving agents to a neighbor pixel with the highest intensity.
Despite the promising results, this idea does not extract all the richness of textural pattern.
Our approach differs from the original ACrawler model in terms of movement: each agent is not only able to move to the higher altitudes as well as to lower ones.
It allows the model to extract the details present in peaks and valleys of the images.

First, the agents move to higher intensities as the original artificial crawler method.
Thus, the artificial crawlers are performed using this rule and the curve $\varphi_{max}$ is obtained.
Throughout the paper, this rule of movement will be referred as $max$.
We can observe that the original artificial crawler method only models the peaks of a textured image.
To obtain a robust and effective texture representation, we propose to move artificial crawlers toward lower intensities $-$ this rule of movement will be referred throughout the paper as to $min$.
In our approach, artificial crawlers are randomly placed in the image with initial energy $\epsilon$.
Then, the movement step is modified as follows:

$$\forall_{i}: e_{i}^{t} > e_{min}, \rho_{i}^{t+1} = f(\rho_{i}^{t})$$

$$f(\rho) =
 \begin{cases}
 \rho_{i}^{t}, & \text{if (a) is satisfied} \\
 \rho_{min}^{t}, & \text{if (b) is satisfied} \\
 \rho_{m}^{t}, & \text{if (c) is satisfied}
 \end{cases}$$ \\
$(a)$ Agents settle down if the grey level of its 8-neighbors are higher than itself. \\
$(b)$ Agents move to a specific pixel if there exist one of its 8-neighbors ($\rho_{min}^{t}$) with lower intensity \\
$(c)$ If there exist more than one neighbor with lower intensity, an agent moves to the pixel that already was
occupied ($\rho_{m}^{t}$).

The multi-agent systems using the rule of movement $min$ is characterized as the original method by using the number of live agents at each time.
Considering that now we have two rules of movement, the final feature vector of our approach is composed by the concatenation of $\varphi_{max}$ and $\varphi_{min}$ according to:
\begin{equation}
\varphi = [\varphi_{max}, \varphi_{min} ]
\label{eq:finalvector}
\end{equation}

\begin{figure}[!htb]
	\centering
		\subfigure{\includegraphics[width=0.49\textwidth]{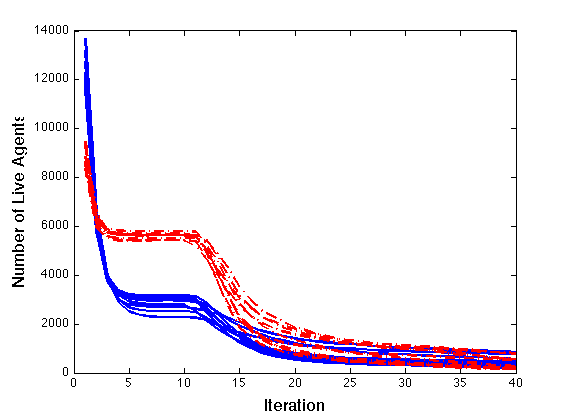}}
		\subfigure{\includegraphics[width=0.49\textwidth]{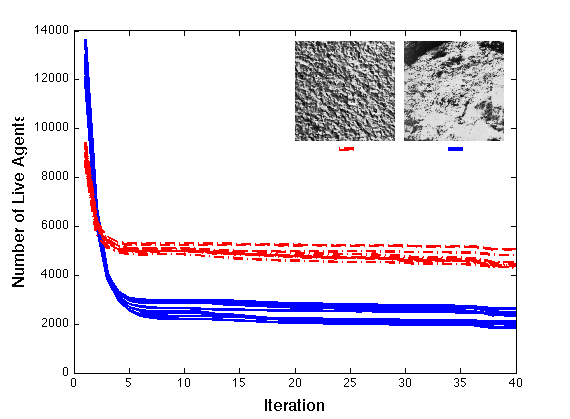}}
	\caption{\label{pic:curve1} Curve of live agents using the rules of movement (left) $max$ and (right) $min$.}
\end{figure}

Figure \ref{pic:curve1} shows the curves of the evolution process.
We took two classes of textures (on the top right-hand corner in Figure \ref{pic:curve1}) from the album the Brodatz \cite{brodatz1966} to illustrate the separability. On the left, Figure \ref{pic:curve1} shows the number of live agents using the rule of movement $min$, while the curve for the rule of movement $max$ is shown on the right of Figure \ref{pic:curve1}.
The experimental results below corroborate the importance of both rules of movement in the texture modeling.

\section{Experimental Results}
\label{sec:exps}

In this section, we demonstrate the effectiveness of our approach. 
We first outline details of the experimental setup, and then, experiments carried out on two datasets are discussed: Brodatz and silk fibroin. Here is described the whole process for image acquisition of silk fibroin scaffolds. Besides, we show comparative results with different texture methods.

\subsection{Experimental Setup}

\begin{figure}[!htbp]
	\centering %
		\includegraphics[width=0.9\textwidth]{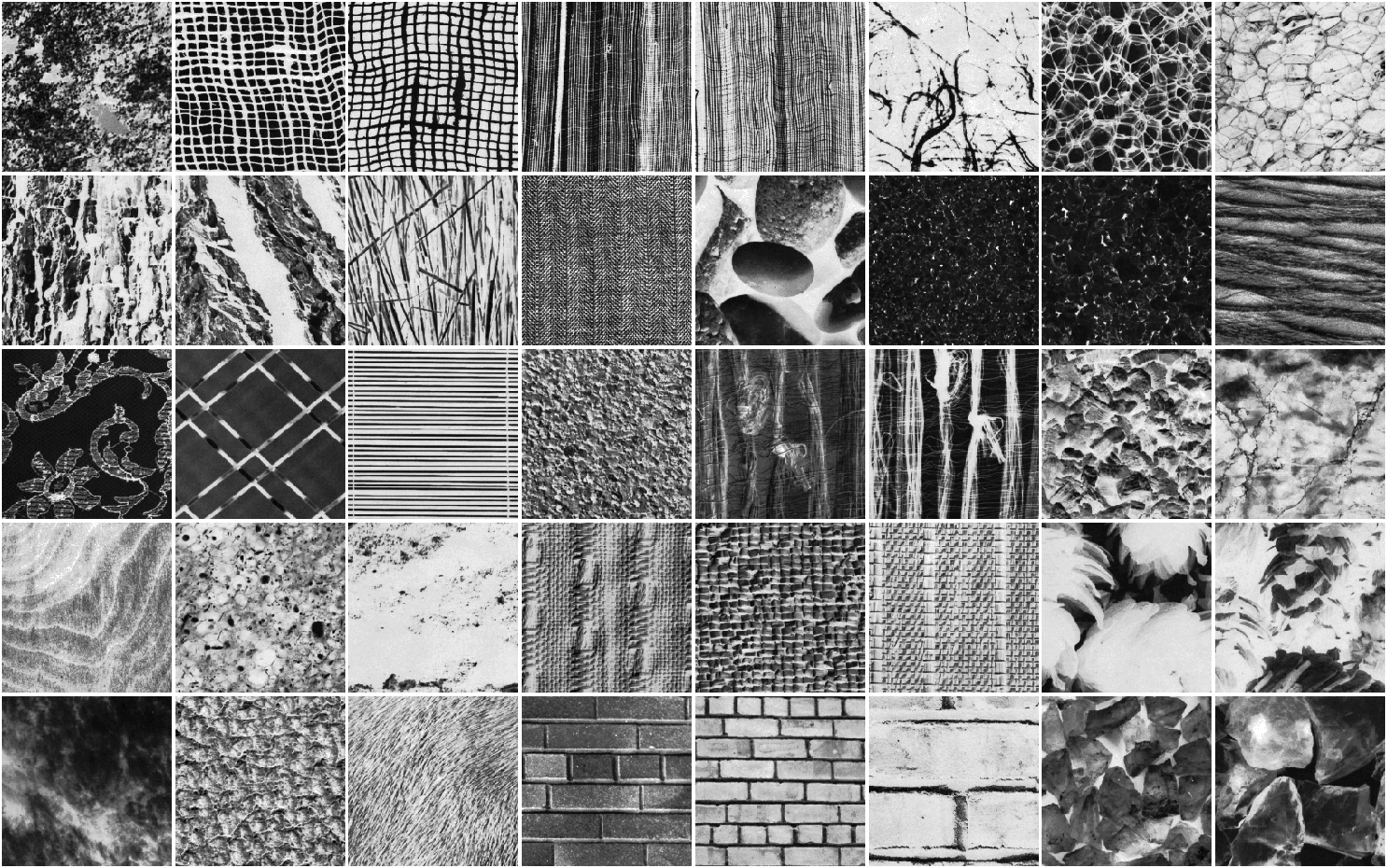}
		\caption{\label{pic:brodatz} Example of 40 Brodatz texture classes used in the experiment. Each image
has $200 \times 200$ pixels and 256 gray levels.}
\end{figure}

The proposed method was first evaluated over texture classification experiments by using images extracted from Brodatz album
\cite{brodatz1966}. This album is considered a well-known benchmark for evaluating texture recognition methods. Each
class is composed by one image divided into nine sub-images non-overlapped. A total of 440 images grouped into 40
classes was considered. Each image has $200 \times 200$ of size and 256 grey levels. One example of each class is shown in Figure \ref{pic:brodatz}.

The texture classification was carried out for ten-fold cross validation to avoid bias.
At each round, we randomly divide the samples of each class into ten subsets of the same size, i.e., nine for training and the remaining for testing. The results are reported as the average value over the ten runs.
For classification, we adopted the model Linear Discriminant Analysis (LDA). The underlying idea is to maximize the Euclidian distance between the means of the classes, while minimizing the within-class variance.
For further information we refer to \cite{fukunaga1990}.

Linear Discriminant Analysis (LDA) \cite{fidlerPAMI2006} was selected since it is well founded in statistical learning theory and has been successfully applied to various object detection tasks in computer vision. LDA, originally proposed by Fisher, computes a linear transformation $(T \in \Re^{d \times n})$ of $D$, which $D \in \Re^{d \times n}$ is a matrix and $d$ denotes the number of features and $n$ number of samples.

We optimized two parameters of the artificial crawler model: the number of agents and the way that agents move in the evolution process.
The number of agents placed on the pixels of the image was inital set to $1000$ with a coverage rate of $10\%$, varying from $1000$ to $35000$.
In our experiments, all agent was born with an initial energy $\epsilon$ of 10 units and the loss for each iteration consumes 1 unit of energy. The absorption rate was set to $0.01$ in terms of the current pixel. For the survival threshold and the upper bound of energy were set to $1$ and $12$ units, respectively.

\subsection{Performance Evaluation}

\begin{figure}[!htb]
	\centering
 		\subfigure[]{\includegraphics[width=0.49\textwidth]{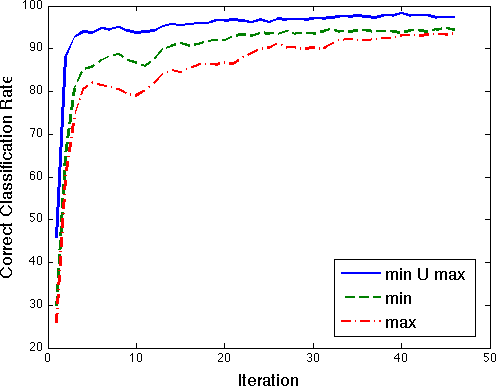}}
		\subfigure[]{\includegraphics[width=0.49\textwidth]{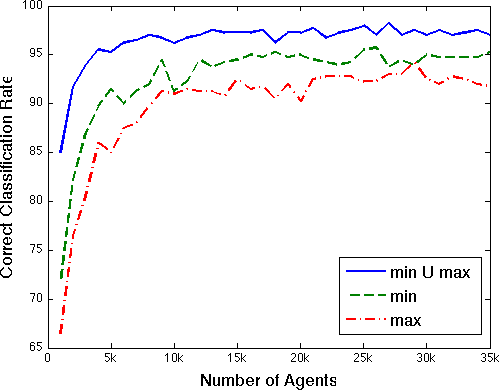}}
		\caption{\label{pic:ccr-vc-numberAgents} Comparison of artificial crawler methods for different values
of (a) iterations and (b) number of agents in the Brodatz dataset.}
\end{figure}

\medskip\noindent\textbf{Experiment 1:} First, we perform an analysis of our method on the Brodatz dataset.
Figure \ref{pic:ccr-vc-numberAgents}(a) presents the correct classification rate versus the number of iterations.
The results for the original artificial crawler is shown as curve $max$ while the results for our method is shown as
curve $min \cup max$.
For a complete comparison, we also provide the results for an method which agents move to pixels with lower intensity
$-$ curve $min$.
As can be seen, the proposed method provided the highest correct classification rates for all values of iterations.
These experimental results indicate that the proposed method significantly improves performance over the traditional
methods.
We can also observe that the rule $min$ provided higher rates than the rule $max$, given the idea that valleys are more
discriminative than peaks in the Brodatz dataset.

Another important parameter of the artificial crawler methods is the number of agents.
Figure \ref{pic:ccr-vc-numberAgents}(b) shows the correct classification rates versus the number of agents.
As in the previous experiment, our method achieved the highest rates compared to the other two strategies.
Again, the rule $min$ provided higher rates than the rule $max$.
Another important observation from Figure \ref{pic:ccr-vc-numberAgents}(b) is that using a few agents, the methods
achieved good classification results, which makes the artificial crawler methods suitable for real time applications.
Using these two plots, we can determine the best parameters of our method to $t_{max}=41$ and $n=27k$.

The results of the proposed method are compared with existing texture analysis methods in Table
\ref{tab:comparisonBrodatz}.
It is observed that the our method outperforms the state-of-the-art.
The highest classification rate of $98.25\%(\pm 1.69)$ was obtained by our method, which is followed by a classification
rate of $95.25\%(\pm 3.43)$ obtained by the Gabor filter, one of the most traditional texture analysis method.

\begin{table}[!htbp]
	\centering
	\small
		\begin{tabular}{|c|c|c|}
			\hline
			Method & Images correctly classif. & Correct classification(\%)\\
			\hline
			Fourier descriptors \cite{azencottPAMI1997} & 346 & 86.50 ($\pm 6.58$) \\
			Co-occurrence matrices \cite{haralickTSMC1973} & 365 & 91.25 ($\pm 2.65$)  \\
			Original artificial crawler \cite{zhangIAT2004} & 372 & 93.00 ($\pm 5.50$) \\
			Gabor filter \cite{gaborJIEE1946} & 381 & 95.25 ($\pm 3.43$)\\
			\textbf{Proposed method} & \textbf{393} & \textbf{98.25 ($\pm 1.69$)} \\
			\hline
		\end{tabular}
	\caption{Experimental results for texture methods in the Brodatz dataset.}
	\label{tab:comparisonBrodatz}
\end{table}

\begin{figure}[!htb]
	\centering 
		\includegraphics[width=0.7\textwidth]{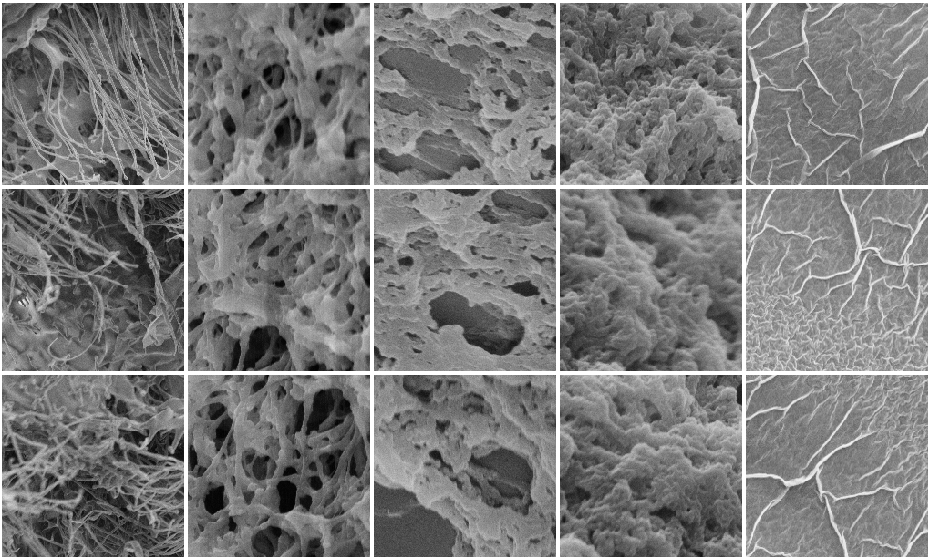}
		\caption{\label{pic:fibroins} Samples for each glycerol concentration. The first column corresponds to
$0\%$ of concentration, the second $2.5\%$, and so on up to $10\%$.}
\end{figure}

\medskip\noindent\textbf{Experiment 2:} In this experiment, we present a comparative study of our approach to assess the quality of the silk fibroin scaffolds. Our goal is to provide an effective method to support the visual analysis, thus reducing the subjectiveness of conclusions based on the human analysis. The potential of the silk fibroin is enhanced by including glycerol solutions during scaffold formation \cite{luBIOMACROMOLECULES2010}. In general, such concentration can range from $0\%$ to $10\%$ with step of $2.5\%$. As far as the authors know, this paper is the first to report a method for characterizing the silk fibroin scaffolds. This dataset contains 5 classes, each of 10 $200\times200$ pixel images. Figure \ref{pic:fibroins} shows three samples for each concentration.

\begin{figure}[!htb]
	\centering
 		\subfigure[]{\includegraphics[width=0.49\textwidth]{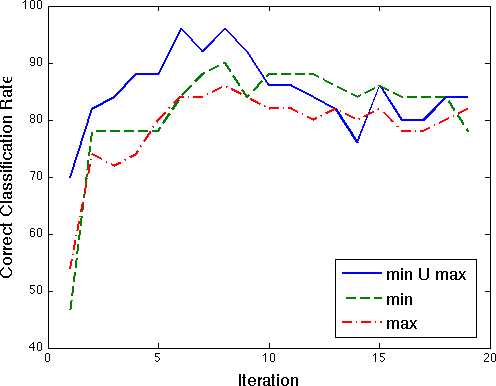}}
		\subfigure[]{\includegraphics[width=0.49\textwidth]{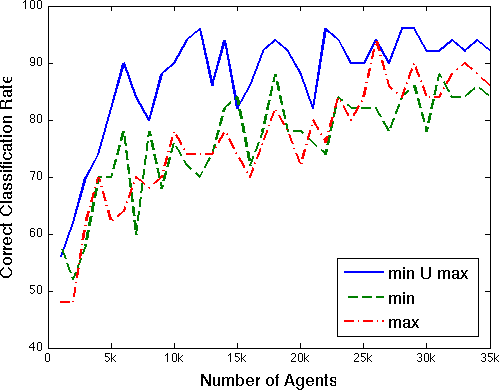}}
		\caption{\label{pic:ccr-vc-numberAgents2} Comparison of artificial crawler methods for different values
of (a) iterations and (b) number of agents in the silk fibroin dataset.}
\end{figure}

We perform the same experiment to determine the best parameters of iteration and number of agents in the Silk Fibroin
dataset.
Figure \ref{pic:ccr-vc-numberAgents2}(a) presents the evaluation of $t$ while Figure \ref{pic:ccr-vc-numberAgents2}(b)
presents the evaluation of $n$ for different artificial crawler methods.
Using both plots, we found that the best results are achieved for $t_{max}=7$ and $n=28k$.

In the silk fibroin dataset, our method achieved highest classification rates when compared with traditional texture
analysis methods.
The experimental results, presented in Table \ref{tab:comparisonFibroina}, shows that our method achieved a
classification rate of $96\%(\pm 8.43)$.
These experimental results indicate that our method is consistent and can be applied in real-world applications.

\begin{table}[!htbp]
	\centering
	\small
		\begin{tabular}{|c|c|c|}		
			\hline
			Method & Images correctly classif. & Correct classification(\%)\\
			\hline
			Fourier descriptors \cite{azencottPAMI1997} & 39 & 78.00 ($\pm 22.01$) \\
			Co-occurrence matrices \cite{haralickTSMC1973} & 47 & 94.00 ($\pm 9.66$)  \\
			Original artificial crawler \cite{zhangIAT2004} & 42 & 84.00 ($\pm 15.78$) \\
			Gabor filter \cite{gaborJIEE1946} & 31 & 62.00 ($\pm 19.44$)\\
			\textbf{Proposed method} & \textbf{48} & \textbf{96.00 ($\pm 8.43$)} \\
			\hline		
		\end{tabular}
	\caption{Experimental results for texture methods in the silk fibroin dataset.}
	\label{tab:comparisonFibroina}
\end{table}

\section{Conclusion}
\label{sec:conclusions}

In this paper we presented a novel approach based on artificial crawler for texture classification. We have demonstrated how the feature vector can be improved by combining \textit{min} and \textit{max} curves, instead of using only the strategy for the maximum of intensity of the pixels. Although traditional methods of texture analysis have provided satisfactory results, the approach proposed here has proved to be superior for characterizing textures. We have tested our proposal on the most popular benchmark for texture analysis and find that it produces good classification results. Furthermore, we tested our approach on the silk fibroin scaffolds analysis and results indicate that our method is consistent and can be applied in real-world applications.

The results support the idea that our approach can be used as a feasible step for many analysis not only applications on tissue engineering. In addition, our results can be improved by studying the setting of our method. We have already implemented other variation. As part of the future work, instead using random sampling of agents we plan focus on evaluating the deterministic sampling, i.e, each pixel of the image is initialized with on agent.

\subsubsection*{Acknowledgments.}
BBM and WNG were supported by FAPESP under grants 2011/02918-0 and 2010/08614-0, respectively. OMB was supported by FAPESP grant 2011/01523-1 and  CNPq grants 306628/2007-4 and 484474/2007-3.
\medskip


\end{document}